\pdfoutput=1
\documentclass[11pt]{article}

\usepackage[]{acl}

\usepackage{times}
\usepackage{latexsym}

\usepackage[T1]{fontenc}

\usepackage[utf8]{inputenc}

\usepackage{microtype}
\usepackage{mathtools}
\usepackage{makecell,multirow}
\usepackage{booktabs}
\title{A Numerical Reasoning Question Answering System  with Fine-grained Retriever and the Ensemble of Multiple Generators for \textsc{Fin}QA }

\author{Bin Wang$^{ab}\footnotemark[1]$ \quad Jiangzhou Ju$^{ab} \footnotemark[1]$ \quad Yunlin Mao$^{ab}$\quad  \\
\textbf{Xin-Yu Dai$^{ab}\footnotemark[2]$\quad Shujian Huang$^{ab}$\quad Jiajun Chen$^{ab}$}\\
$^{a}$National Key Laboratory for Novel Software Technology, Nanjing University \\ 
$^{b}$Collaborative Innovation Center of Novel Software Technology and Industrialization, Nanjing \\ 
\texttt{\{wangbin, jujiangzhou,maoyl\}@smail.nju.edu.cn} \\
\texttt{\{daixinyu, huangsj,chenjj\}@nju.edu.cn}
}

\begin{document}
\renewcommand{\thefootnote}{\fnsymbol{footnote}}
\maketitle
\begin{abstract}
The numerical reasoning in the financial domain---performing quantitative analysis and summarizing the information from financial reports---can greatly increase business efficiency and reduce costs of billions of dollars. Here, we propose a numerical reasoning question answering system to answer numerical reasoning questions among financial text and table data sources,  consisting of a retriever module, a generator module, and an ensemble module. Specifically, in the retriever module, in addition to retrieving the whole row data, we innovatively design a cell retriever that retrieves the gold cells to avoid bringing unrelated and similar cells in the same row to the inputs of the generator module. In the generator module,  we utilize multiple generators to produce programs, which are operation steps to answer the question. Finally, in the ensemble module, we integrate multiple programs to choose the best program as the output of our system. In the final private test set in FinQA Competition, our system obtains 69.79 execution accuracy.

\end{abstract}
\footnotetext[1]{Authors contributed equally}
\footnotetext[2]{Corresponding author}
\renewcommand{\thefootnote}{\arabic{footnote}}
\setcounter{footnote}{0}
\section{Introduction}
Financial analysis---performing quantitative analysis and summarizing the information from financial reports---is important for business companies to make appropriate investments and decisions; otherwise, terrible analysis can cause unpredictable cost\citep{mackenzie2008engine}. However, with the explosive growth of electronic financial documents, the sufficient quality demand for the financial analysis can't be met by dedicated human effort.

Hence, ~\citet{chen2021finqa} introduced \textsc{FinQA}, a financial question answering~(QA) dataset,  which requires complex numerical reasoning ability, and hoped the researchers could devise  models to automate to analyze financial data deeply. Answering questions in \textsc{FinQA} involves ten common calculations, such as addition, multiplication, and table aggregation operations, among text and table data sources. In addition to outputting the answer, the models also need to generate executable reasoning programs. So devising models to meet the numerical reasoning ability in \textsc{FinQA} is challenging.

The works in \textsc{FinQA} mainly utilize a retriever-generator QA framework, which first acquire supporting facts from financial documents and then generate executable reasoning programs to obtain the program to the question. \citet{chen2021finqa} proposed a row retriever to retrieve the rows as support facts and a pre-trained language model based generator to generate the program.

However, in the retrieval phase, the row retriever in the table will introduce noises to inputs of the generator. 
For the table data, the meaning of the cell value consists of three parts: the row name, the column header, and the cell value, so the cells in the same row are very similar. The row retriever retrieves a row including a gold cell as inputs of the following generator, which lead to unrelated cells~(noises) from the same row to disturb the generator. Meanwhile, the questions in \textsc{FinQA} requiring numerical reasoning are very complex, so the single generator is not competent for such a task. 

Considering these challenges, we provide a numerical reasoning system with the cell retriever to retrieve the gold cells---the cells both occur in annotated programs and annotated table row---instead of the whole row data and the ensemble of multiple generators to solve complex questions in \textsc{FinQA} dataset. 
Specifically, we first innovatively devise a cell retriever module which classifies whether the cell is support fact compared with the row retriever. In the generator module, 
we feed the retrieval results of the cell retriever and the row retriever to both the generator of FinQANet and Unified pre-trained Language Model~(UniLM)\citep{dong2019unified} respectively to obtain four candidate programs.
In the ensemble module, we use heuristic rules to integrate candidates to get the final program.


The main contribution of our work can be concluded as follows:
\begin{itemize}
    \item  We first innovatively devise a cell retriever module to retrieve gold cells to avoid bringing unrelated cells in the same row to the inputs of the generator module.
    \item We integrate multiple generators to improve the performance of our system to generate the program.
    \item Our QA system obtains 69.79 execution accuracy  in the final private test set in FinQA Competition.
\end{itemize}

\section{System Description}
\begin{figure*}
    \centering
    \includegraphics[width=16cm]{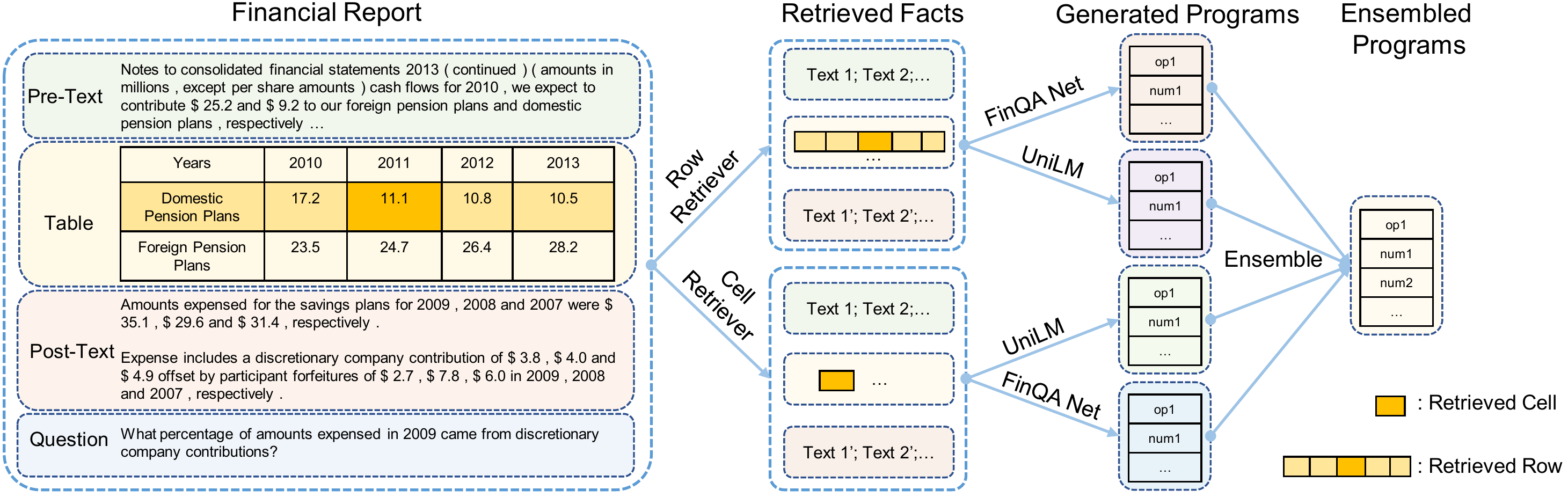}
    \caption{The architecture of our system. The retriever module firstly respectively retrieve cells and rows as the model input of generators, then multiple generators reproduce different programs, finally, the ensemble module integrates the all programs to decide the final program.}
    \label{fig:structure}
\end{figure*}
As Figure~\ref{fig:structure} shows, our model consist of three modules: the retriever module, the generator module, the ensemble module. In the following, we will respectively introduce these modules.
\subsection{Task Definition}

Given a financial report consists of text and table information, and a numerical reasoning question with respect to the input financial report, the task is to generate the correct reasoning programs, which can be executed to get the correct numerical answer.

\subsection{Retriever Module}


In this section, we mainly introduce our  fine-grained retriever and our retriever module includes the row retriever of~\citet{chen2021finqa}. 

Retriever is a part of our system, which retrieves supporting facts from the input financial report. As the questions in the \textsc{FinQA} dataset have complex background information which consists of two forms of text and table and generally exceeds 512 tokens, which makes it difficult to put the whole example into a pre-trained language model like BERT \citep{devlin-etal-2019-bert} without any preprocessing. So we should retrieve parts of the text and the table data. Furthermore,
according to the statistics, 76.58\% of the questions require the table information to get the answer, so how to retrieve the table data is very important.

Here, we propose a cell retriever, which retrieves essential cell values instead of the whole row data. The row retrieving method (Row Retriever) will bring unrelated cell data into facts, and these cell data are highly similar, which makes it hard for the generator to choose the right number. In contrast, our method (Cell Retriever) avoids this problem at the retrieval stage. The effectiveness of using the cell retriever is also confirmed by experiments in section \ref{subsection:retriever_res}.

Specifically,  to convert the cell data into text format, we get the gold cell by finding the single cell which contains the number that appears in the gold program, with a heuristic method. After converting table information into text format, we got the total text facts. Text facts containing numbers that appear in the gold program are called positive fact, the rest is called negative fact. 
With all the positive and negative text facts, we trained a retriever to retrieve related facts, which serve as the input of the generator. Since the number of negative facts is much more than positive facts, following \citet{chen2021finqa}, we performed negative sampling during training to ensure that the ratio of positives and negatives in each batch is 1:3.

We also tried the clustering negative sampling method, hoping to improve the retrieval performance by mixing a small number of high-quality negative examples and common negative examples. However, in the FinQA dataset, the ratio of positive and negative cases is about 1:20, which makes it possible to use all data. And the result of this method is worse than that of using random sampling of all data. 
\subsection{Generator Module} 
After the retrieval, the question text is combined with the retrieved supporting facts as input to the generator, which then outputs an executable program that answers the question. 

We use two end-to-end generative models, the generator used in FinQANet \citep{chen2021finqa} and the Unified pre-trained Language Model~(UniLM)~\citep{DBLP:conf/nips/00040WWLWGZH19}, to model the generation process from the input text to the output program. The effectiveness of the generator of FinQANet has been verified in \citet{chen2021finqa} by many comparative experiments, and Unilm also has achieved superior performance on Natural language understanding and generation tasks. Herein, we use these two models as generators for our system. The former first encodes the input text with pre-trained LMs and obtains the output embeddings, and then uses an LSTM to decode to get the final program. The modeling of the latter is achieved by employing a shared network and utilizing specific self-attention masks to control what context the prediction conditions on, which enables sequence-to-sequence generation without the need for an additional decoder.

These two generative models have the same input format, but there is a slight difference in the output sequence. In UniLM, we use the vocabulary of the pre-trained language model as the vocabulary of the decoder. We first tokenize the target program into tokens and then join them into a sequence using special symbols (e.g., \$) as separators. In addition, to ensure that UniLM outputs the correct executable program, we use heuristics such as edit distance to correct the operators that involve table operations, and beam search in the inference stage to further improve the correctness of the generated programs.

\subsection{Ensemble Module}
In this section, we will describe how our system ensembles multiple models. The methods used are specified as follows.
\paragraph{Checkpoints Average}
After each training epoch, a new saved model is obtained. Based on the performance of these models on the validation set, a number of the best-performing ones are selected for parameter averaging to obtain the final model. We applied this method to the retrieval and generation models, respectively, and the experimental results show that the method can steadily improve the performance. 
\paragraph{Loss Ensemble}
For the generator in FinQANet, we can train two generators with the input retrieved based on cell and row, respectively. Each generated program corresponds to a loss output, and a lower loss indicates that the model fits the data better and the probability of correct output is higher. When integrating the outputs of the two generators, we prefer the one with a more minor loss. 
\paragraph{Score Ensemble}
For the UniLM, we can also train two generators with the input retrieved based on cell and row, respectively. A probability score output can be obtained while generating a program by beam search, and a higher score indicates that the probability of correct output is higher. When integrating the outputs of the two generators, we prefer the one with a higher score. 
\paragraph{Mixed Ensemble}
Using two different patterns of retrieval and two generators, we can get four outputs: $\{O_{cf}, O_{cu}, O_{rf}, O_{ru}\}$, where $c$ and $r$ denote cell and row-based retrieval, respectively, and $f$ and $u$ denote FinQANet's generator and UniLM, respectively.
We first combine the two UniLM-based outputs ($O_{cu}$ and $O_{ru}$) into $O_{u}$ using the score ensemble, and then set two thresholds: $T_{loss}$ and $T_{score}$. 1) when $O_{cf}$'s loss is lower than that of $O_{rf}$, if $O_{cf}$ is a non-executable program or its corresponding loss is greater than $T_{loss}$ and $O_{u}$'s score is greater than $T_{score}$, we choose $O_{u}$ as the final output, otherwise we take $O_{cf}$ as the final output. 2) when $O_{cf}$'s loss is equal or higher than that of $O_{rf}$, if $O_{rf}$ is a non-executable program or its corresponding loss is greater than $T_{loss}$ and $O_{u}$'s score is greater than $T_{score}$, we choose $O_{u}$ as the final output, otherwise we take $O_{rf}$ as the final output. In our system, $T_{loss}$ and $T_{score}$ are set to $0.01$ and $-0.15$, respectively.

\section{Experiment}
In this section, we mainly introduce the experimental effect of our reasoning QA system and show the advantages of our system from an experimental comparison view. The reported performance of all single models is based on the average of the best checkpoints.
\subsection{Data Usage Introduction}
We only utilize the data from the FinQA dataset which includes the training set 6251, validation set 883, and public test set 1147. Specifically, for the public test set, the results of the validation set are used to select the model parameters.

\subsection{Model Setting \label{sec:model_setting}}
\paragraph{Retriever}

We use the same model architecture and parameter setting for both row retriever and cell retriever. We use BERT-base with the maximum sequence length of 512 as the classifier, Adam optimizer \citep{kingma2014adam} with learning rate of 2e-5, batch size of 24, and epochs of 20. We train each model on a single NVIDIA GeForce RTX 3090 GPU.  After training, we perform model parameter averaging on the 5 best checkpoints to get a more stable result. Finally, for Row Retriever, the top 3 retrieved facts serve as input of the generator, while the top 5 for Cell Retriever.

\paragraph{Generator}
For the generator of FinQANet, we use RoBERTa-large with the maximum sequence length of 512 as the encoder, Adam optimizer \citep{kingma2014adam} with learning rate of 1.5e-5, batch size of 28, and epochs of 300. We train the model on 4, 32GB NVIDIA V100 GPUs. For Unilm, we use RoBERTa-large with the maximum sequence length of 512 as the encoder, Adam optimizer  with learning rate of 1.5e-5, batch size of 12, epochs of 100, beam search size of 3. We train the Unilm on one 32GB NVIDIA V100 GPU.

\subsection{Retriever Module Resutls}
\label{subsection:retriever_res}

\begin{table}[]
\centering
\begin{tabular}{l|c} \toprule
                        & \multicolumn{1}{c}{Test Acc} \\ \midrule
Row Retriever~(top 3)                          & 89.06                          \\
Row Retriever~(top 5)                              & 93.48                             \\
Cell Retriever~(top 3)                          & 85.02                          \\
Cell Retriever~(top 5)                            & 90.12                             \\ \bottomrule
\end{tabular}
\caption{The recall rate of retrieval models on validation set and public test set}
\label{tab: cell_retrieval}
\end{table}

\begin{figure}[!h]
    \centering
    \includegraphics[width=7.5cm]{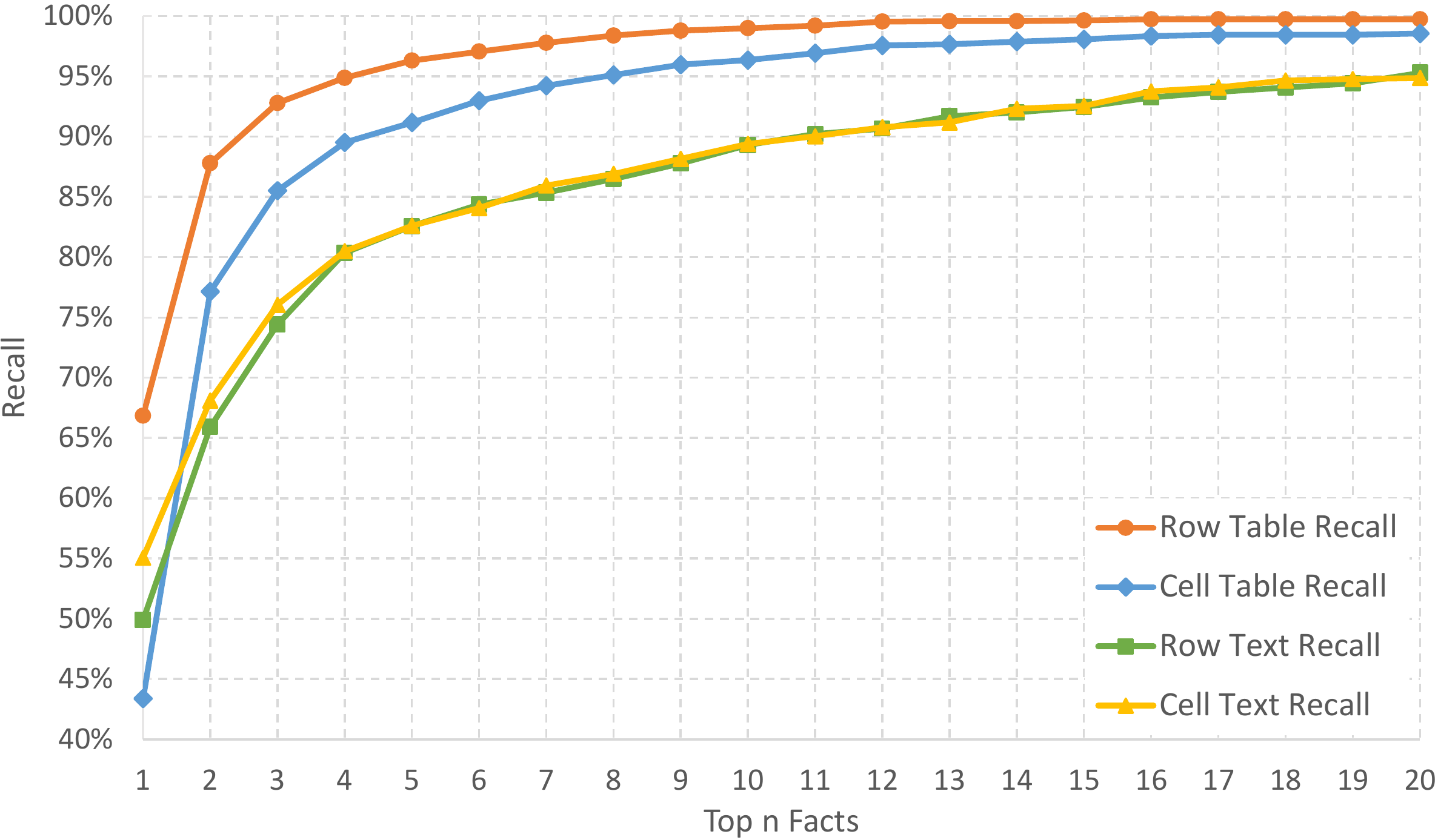}
    \caption{Recall rate of row retriever and cell retriever in text and table in top $n$ facts} 
    \label{fig:topn_recall}
\end{figure}

In this section, we compare the recall rate of the cell retriever with the row retriever.
\paragraph{Overall Recall Rate}  Here, the recall rate of the row retriever indicates the ratio of retrieved rows to the gold rows and the recall of the cell retriever indicates the proportion of the retrieved cells to the gold cells. From Table~\ref{tab: cell_retrieval}, the cell retriever is 3.95 accuracies lower than the row retriever in the top 3 recall in the test set; The gap between these two retrievers is reduced to 2.21 in the top 5 recall, which shows the gap will become smaller with the increase of the numbers of retrieved facts. However, as the template of a single cell is shorter, we can select the top 5 facts in the cell retriever as the inputs of the following generators under the max length limit of the pre-trained model. So the recall~(90.12 in top 5) of the retrieved facts in the cell retriever is higher than the recall~(89.06 in top 3) of the row retriever.



\paragraph{Table and Text Recall Rate}
We further analyze the recall rate of the  retrievers among the table and text after selecting the top n facts based on the scores from the corresponding retrievers. From the Figure~\ref{fig:topn_recall}, we can see the performance of the cell retriever in the text is better than the row retriever; the performance of the cell retriever in the table is the opposite. The poor performance of the cell retriever in the table can be reduced to that the retrieval of  gold cells are more easily disturbed by non-gold and unrelated cells in the same row. However, the template length of single cell fact is shorter and we can select more top n facts. For the cell retriever, we select the top 5 facts as the generator inputs. In Figure~\ref{fig:topn_recall}, we can see the table recall of the cell retriever in the top 5~(91.16) is close to the row retriever in the top 3~(92.78).  

\subsection{Generator and Ensemble module Results}
In this subsection, we respectively introduce the performance of the generator module and the ensemble module.
\paragraph{Generator Module} 
We first show the performances of the multiple single generators in the public test set. In this subsection, we refer FinQANet as its generator. Our multiple generators respectively include the FinQANet~(Row Retriever), FinQANet~(Cell Retriever), and Unilm~(Row Retriever), Unilm~(Cell Retriever).
The FinQANet~(Row Retriever) and Unilm~(Row Retriever) are both trained on the facts from the Row Retriever.  The FinQANet~(Cell Retriever) and Unilm~(Cell Retriever)  are both trained on the facts from the Cell Retriever.
As seen in Table~\ref{tab:qa_pub} upper part, FinQANet~(Row Retriever) achieves the best performance among all the single models. 
As most of the questions in \textsc{FinQA} require the table information and the recall rate of the cell retriever is lower than the row retriever, this may cause the performance of the generators based on the cell retriever to be poor. 
In addition, compared the cell retriever using  gold cells with the row retriever using gold rows, the cell retriever surpasses 5.43 execution accuracy, which shows the potential of the cell retriever to enhance the generator.

\paragraph{Ensemble Module} 
In Table~\ref{tab:qa_pub} bottom part, the results of different ensemble strategies are shown. FinQANet~(Row and Cell Retriever) is the ensemble of the generators whose inputs are from the cell retriever and the row retriever respectively. FinQANet~(Row and Cell Retriever) is 2.44 execution accuracy higher than the single model, which shows the effectiveness of the loss ensemble. Unilm~(Row and Cell Retriever) is the ensemble of the two Unlim whose inputs are from the cell retriever and the row retriever, whose performance also can prove the score ensemble is effective. FinQANet+Unilm (Row Retrieve) is the ensemble of the FinQANet's generator and Unilm whose inputs are all from the row retriever, whose performance shows the cell retriever brings differentiated contribution for the generator. Finally, our system obtain the best performance 68.00 execution accuracy in the test set, which proves the effectiveness of the mixed ensemble.

Table~\ref{tab: qa private} show the results of our system in private results. Without adding validate set and public test set of \textsc{FinQA}, our system can achieve 66.92 execution accuracy.  Our system finally achieves 69.97 execution accuracy on the private test set after using validate set and public test set.

\begin{table}[]
\centering
\small
\begin{tabular}{l|ll} \toprule
                               & Exe Acc & Prog Acc \\ \midrule
FinQANet~(Row Retriever)       & 63.90  & 61.98   \\
FinQANet~(Cell Retriever)      & 62.94   & 60.33    \\
Unilm~(Row Retriever)          & 60.76   & 58.76    \\
Unilm~(Cell Retriever)         & 60.50   & 57.97    \\ \midrule
FinQANet~(Gold Row)       & 70.00  & 68.76   \\  
FinQANet~(Gold cell)       & 75.43  & 73.71   \\  \midrule
FinQANet~(Row and Cell Retriever) & 66.34  & 64.08 \\ 
Unilm~(Row and Cell Retriever)  & 63.93  & 61.46 \\
FinQANet+Unilm (Row Retriever) & 66.08   & 63.99    \\
Our system                      & \textbf{68.00}   & \textbf{65.21}    \\ \bottomrule
\end{tabular}
\caption{The execution accuracy (Exe Acc) and program accuracy (Prog Acc) for the multiple single models and ensemble models on public test set. FinQANet~(Gold cell) indicates the inputs of the corresponding generator are gold cells. }
\label{tab:qa_pub}
\end{table}

\begin{table}[!tbp]
\centering
\small
\begin{tabular}{l|ll}
\toprule
                         & \multicolumn{1}{c}{Exe Acc} & \multicolumn{1}{c}{Prog acc} \\ \midrule
FinQANet$^{\star}$~(RoBERTa-large) & 59.70                       & 54.95                        \\
Our system~(w / dev+test)                & 66.92                          & 62.46                          \\ 
Our system                & 69.97                          & 65.40                          \\ \bottomrule
\end{tabular}
\caption{The execution accuracy (Exe Acc) and program accuracy (Prog Acc) for our system on private test set. FinQANet$^{\star}$~(RoBERTa-large) is the approach of \citet{chen2021finqa} }
\label{tab: qa private}
\end{table}

\section{Related work}
\paragraph{Numerical Reasoning QA}
Currently, the Math Word Problem~(MWP) involves a lot of studies of numerical reasoning.
Popular datasets includes Math23K~\citep{wang-etal-2017-deep}, Ape210K~\citep{zhao2020ape210k}, MathQA~\citep{amini2019mathqa}, MAWPS~\citep{koncel2016mawps} which almost annotated gold math expression. As the math expression can be regarded as either a sequence or a structured tree, researchers explored many neural architectures: sequence-to-sequence,  sequence-to-Tree, and transformer-to-tree. Furthermore, \citet{zhang2020graph} use the graphs to capture the relations among numerical quantities, their description, and unknown quantities. Compared with the methods in MWP, we select FinQANet as our one generator, which utilizes sequence-to-sequence to solve problems. In addition, for reading comprehension, the dataset DROP also includes some simple calculations over texts. The mainstream approaches in DROP always use multiple heads for predicting spans, the kind of calculation. However, this kind of approach is hardly transferred to FinQA dataset.

\paragraph{Table QA}
Many Table Question Answering works like TAPAS~\citep{herzig2020tapas}, GraPPA~\citep{yu2020grappa}, use pre-trained Tabular Language Models (TaLMs) such as TaBERT~\citep{yang-etal-2022-tableformer}, TUTA~\citep{wang2021tuta}, TURL~\citep{deng2022turl}, and TAPAS to enhance the performance of table question answering. These TaLMS always linearize raw tables row by row, use well-designed encoders to capture the structural information of tables, and are trained by self-supervised tasks and some task-specific objectives. TAPAS extended BERT's architecture to encode linearized raw tables and solve Table QA by predicting the denotation. GraPPa utilized TaLMs to predict whether columns and operations appear in the SQL query. In this paper, our multiple generators take the linearized rows from the template without considering the structural information of the tables. How to apply TaLMs to our generators should be taken into consideration.

\section{Conclusion}
This paper introduces a numerical reasoning QA system with a novelty cell retriever and the ensemble of multiple generators. In the retriever module, the cell retriever sacrifices acceptable accuracies to reduce the unrelated cells into support facts for generators. In the generator module, the ensemble of the multiple generators gains an additional approximately 2.5 execution accuracy improvement compared with the best single generator. However, further exploration is required in the following directions. Due to poor performance of the cell retriever,  the advantages of the cell retriever are not clearly reflected in the generators. So we should improve the recall of the cell retriever. Moreover, how to apply TaLMs into our generators should be taken into consideration. Finally, our ensemble is heuristic and more sophisticated methods should be tried.

\bibliography{anthology,custom}
\bibliographystyle{acl_natbib}

\end{document}